\documentclass[runningheads]{llncs}
\usepackage{graphicx}
\usepackage{subfigure}
\usepackage{color}
\usepackage{booktabs}
\usepackage{multirow}
\usepackage[square,comma,numbers,sort&compress,sectionbib]{natbib}
\usepackage{tabularx}

\usepackage[table]{xcolor}
\usepackage{url}
\usepackage{hyperref}

\begin{document}
\title{Semantic Answer Type Prediction using BERT\thanks{Copyright \textcopyright\ 2020 for this paper by its authors. Use permitted under Creative Commons License Attribution 4.0~International (CC~BY~4.0). {SeMantic AnsweR Type prediction task (SMART) at ISWC 2020 Semantic Web Challenge}}}
\subtitle{IAI at the ISWC SMART Task 2020}
%
%
\author{Vinay Setty \and
Krisztian Balog
%
\authorrunning{V. Setty et al.}
%
\institute{University of Stavanger, Norway}\\
\email{\{vinay.j.setty,krisztian.balog\}@uis.no}}
\maketitle              
\begin{abstract}
This paper summarizes our participation in the SMART Task of the ISWC 2020 Challenge.  A particular question we are interested in answering is how well neural methods, and specifically transformer models, such as BERT, perform on the answer type prediction task compared to traditional approaches.  Our main finding is that coarse-grained answer types can be identified effectively with standard text classification methods, with over 95\% accuracy, and BERT can bring only marginal improvements.  For fine-grained type detection, on the other hand, BERT clearly outperforms previous retrieval-based approaches.  

\keywords{Answer type prediction \and answer category classification \and natural language understanding \and question answering.}
\end{abstract}

\section{Introduction}

The importance of being able to identify the types or semantic categories of answers requested has been long recognized in question answering (QA) research as a key step towards interpreting the meaning of natural language questions~\citep{Ferrucci:2010:BWO,shen2019multi}.
This task may be performed either against a set of coarse-grained types (e.g., at the TREC QA track~\citep{Voorhees:2001:TQA}) or against fine-grained type systems of knowledge bases, such as DBpedia~\citep{Balog:2012:HTT,Garigliotti:2017:TTI}.
The Semantic Answer type prediction (SMART) task~\citep{smart2020}\footnote{\url{https://smart-task.github.io}}, organized as a challenge at the 2020 International Semantic Web Conference (ISWC '20) provides a large-scale evaluation platform for assessing answer type prediction both at course-grained and fine-grained taxonomical levels.  

Specifically, given a natural language question as input, first a high-level answer type category is to be predicted, which can be one of \texttt{resource}, \texttt{literal}, or \texttt{boolean}. 
If the predicted category is \texttt{resource}, a more specific ontological class is to be provided, using the type system of DBpedia or Wikidata.  If the predicted category is \texttt{literal}, it also has to be further classified as \texttt{number}, \texttt{date}, or \texttt{string}.  
In this paper, we refer to the task of coarse-grained answer detection as \emph{category classification} and to the problem of fine-grained prediction of (resource) types as \emph{type prediction}.
Table~\ref{tab:examples} shows some examples.
As seen from the examples, answers for the resource category are provided as a ranked list of types.

\begin{table}[t]
    \caption{Example questions with respective answer categories and types (resources are based on DBpedia types).}
    \label{tab:examples}
    \centering
    \begin{tabular}{|l|l|p{2.5cm}|}
    \hline
    \textbf{Question} & \textbf{Category} & \textbf{Type} \\
    \hline
    \hline
    Who are the gymnasts coached by Amanda Reddin?
    & \texttt{resource}
    & \texttt{["dbo:Gymnast", "dbo:Athlete", "dbo:Person", "dbo:Agent"]} \\
    \hline
    How many superpowers does wonder woman have?
    & \texttt{literal}
    & \texttt{["number"]} \\
    \hline
    When did Margaret Mead marry Gregory Bateson?
    & \texttt{literal}
    & \texttt{["date"]} \\
    \hline
    Is Azerbaijan a member of European Go Federation?
    & \texttt{boolean}
    & \texttt{["boolean"]} \\
    \hline
    \end{tabular}
\end{table}

The main research objectives in this work are to assess (1) How do neural approaches perform compared to  traditional feature-based classification approaches on the category classification task? 
(2) How do neural classification approaches fare against well-establised (fusion-based) IR approaches on the type prediction problem?
We find that (1) is essentially a ``solved'' problem.  Our baseline SVM classifier with word unigrams as features achieved 95\% accuracy.  Neural approaches yield only minor improvements.
As for (2), type prediction has previously been approached as a ranking problem, due to the large number of possible types ($\sim$760 types in DBpedia and $\sim$50k types in Wikidata) that rendered classification-based approaches infeasible. We draw on recent work on extreme multi-class classification and demonstrate substantial gains over the IR baselines. 
It appears that fine-grained type detection on Wikidata is more challenging than on DBpedia.  However, the two are not directly comparable due to the different evaluation measures that are employed, which calls for further analysis.

Code and resources developed in this work are made publicly available at \url{https://github.com/iai-group/smart-task}.

\section{Approach}
\label{sec:approach}

We follow a two-phase approach. In the first phase, we perform \emph{category classification}, that is, a supervised classifier predicts the high-level category of the answer type.  Then, in the second phase, we perform \emph{type prediction} to identify the top-$k$ types for the questions for which answer type was predicted to be a resource.  
For category classification we use two classifiers: SVM with word unigrams as features and fine-tuned BERT (Section~\ref{sec:approach:phase1}).
Type prediction has previously been approached as a ranking task~\citep{Balog:2012:HTT,Garigliotti:2017:TTI}, due to the large number of possible types.  As an alternative, we can cast it as an extreme classification problem (Section~\ref{sec:approach:phase2}).

\subsection{Category Classification}
\label{sec:approach:phase1}
 We flatten the high-level categories into following five categories:~\texttt{boolean},  \texttt{lite\-ral-number}, \texttt{literal-string}, \texttt{literal-date}, and \texttt{resource}.  Since the category classification task is same for both DBpedia and Wikidata, we combine the training datasets for the two and predict the categories for their respective test datasets using the combined model.

\subsubsection{Feature-based classification}

As a first approach to category classification, we use TF-IDF-weighted word unigrams as features. The vocabulary construction and IDF computations are based only on the training portion of the dataset, to avoid any assumptions on the test data. 
Our implementation is based on the \texttt{CountVectorizer} and \texttt{TFIDFVectorizer} classes from the \texttt{sklearn} library\footnote{\url{https://scikit-learn.org/}} with default parameters. We then train an SVM classifier with a linear kernel. We also experimented with using a Naive Bayes classifier, but decided to exclude that after observing inferior performance.

\subsubsection{Neural approach} 
As a second approach, we fine-tune a pre-trained BERT model (RoBERTa)~\citep{Liu:2019:ROB} for a sequence classification task to classify the category.  Our implementation uses the HuggingFace API\footnote{\url{https://huggingface.co/}} for fine-tuning and category classification.

\subsection{Type Prediction}
\label{sec:approach:phase2}

\subsubsection{IR-based methods}
We employ two ranking-based approaches from~\citep{Balog:2012:HTT}, which were introduced for the task of identifying target types of (entity-bearing) search queries.  These approaches are representatives of two main families of object ranking strategies, which have been termed as early and late fusion design patterns in~\citep{Zhang:2017:DPF}.
According to the \emph{type-centric} (TC, a.k.a. early fusion) approach, first a textual representation is built for each type by concatenating the descriptions of entities that are assigned that type.  Then, these type description (pseudo) document can be ranked using standard IR models.  Specifically, we use the DBpedia short abstracts of entities and then rank type documents using BM25.
The second strategy is termed \emph{entity-centric} (EC, a.k.a. late fusion).  There, the top-$k$ most relevant entities from the underlying knowledge base are retrieved using the question as a query.  Then, the relevance score of a given type is computed by aggregating the relevance scores of entities with that type. 
We use BM25 as the underlying retrieval model and a ``catch-all'' entity representation, following the settings in~\citep{Garigliotti:2017:TTI}.  The cut-off parameter $k$ is chosen empirically based on the training set ($k=20$).

\subsubsection{Neural method}

Due to the large number of possible labels, using standard Transformer models is not feasible.  Instead, we cast the type prediction task as an extreme multi-label text classification (XMC) problem: given a question as input text, return the top-$k$ most relevant types from a large collection of possible types.  Vanilla transformer models such as BERT~\citep{Devlin:2019:BER}, RoBERTa~\citep{Liu:2019:ROB}, and XLNet~\citep{Yang:2019:XLN} are ineffective in this scenario due to the memory and computation requirements imposed by the large number of possible labels. This was also confirmed from our experiments that fine-tuning the above mentioned transformer models using the Huggingface framework exhausted all the memory on a 32GB Nvidia Tesla V100 GPU. While this may work on a GPU with larger memory, since we do not have access such a GPU we could not verify and it may still be computationally very expensive to train them.  In addition to the computational limitations, as we show in Section~\ref{sec:analysis}, the types are very sparse with most of them having only a few training instances. In order to address these challenges, a model designed for XMC is essential.  We use the recent solution to extend the transformer models for XMC  coined \emph{X-Transformers}~\citep{Chang:2020:TAM} for this purpose, which shall be referred to as \emph{XBERT} in the rest of the paper.

XBERT consists of three components:
\begin{enumerate}
    \item  Semantic Label Indexing (SLI), which performs hierarchical clustering on the labels to reduce the label space.
    \item Deep Neural Matching (DNM), to fine-tune the Transformer models for each of the label clusters identified by SLI.
    \item Ensemble Ranking (ER), which ranks the instances within the label clusters by training a linear ranker conditionally on the label clusters and the DNM Transformer’s output. 
\end{enumerate}

\section{Experimental Evaluation}

In this section, we discuss our experimental setup, introduce the evaluation measures, and present our results.

\subsection{Data}

Table~\ref{tab:dataset} presents descriptive statistics for the two datasets, DBpedia and Wikidata.  We notice that Wikidata has slightly more \texttt{resource} than \texttt{literal} answer types, compared to DBpedia. 

\begin{table*}[t!]
\caption{Dataset statistics.}
\label{tab:dataset}
\centering
\begin{tabular}{l@{~~}r@{~~}r@{~~}r@{~~}r@{~~}r@{~~}r@{~~}r} 
    \toprule
    \multirow{2}{*}{\textbf{Dataset}} & & \multicolumn{2}{c}{\textbf{Questions}} & & \multicolumn{3}{c}{\textbf{Categories}} \\
    \cline{3-4} \cline{6-8}
    & & \textbf{Train} & \textbf{Test} & & \textbf{Boolean} & \textbf{Literal} & \textbf{Resource} \\
    \midrule
    DBpedia & & 17,571 & 4,393 & & 2,799 & 5,188 & 9,584\\
    Wikidata & & 18,251 & 4,571 & & 2,139 & 4,429 & 11,683\\
    \bottomrule
\end{tabular}

\end{table*}

\subsection{Methods}

The following methods are compared:
\begin{itemize}
    \item SVM: Support Vector Machine for category classification
    \item BERT: RoBERTa for category classification
    \item XBERT: X-Transformers for type prediction
    \item IR/TC: Type-centric IR approach for type prediction  
    \item IR/EC: Entity-centric IR approach for type prediction
\end{itemize}
We train all neural models on a single Nvidia Tesla V100 GPU with 32GB memory.

\subsection{Evaluation Metrics}

\emph{Category classification} is evaluated in terms of classification accuracy. 
\emph{Type prediction} is cast as a ranking task and is evaluated using rank-based metrics.  It, however, considers only those questions that fall into the
\texttt{literal} or \texttt{resource} answer categories.
Furthermore, evaluation is performed differently for DBpedia and for Wikidata, given the nature of their respective type taxonomies.
Types in the DBpedia Ontology are organized hierarchically, up to 7 levels deep.  There, a graded evaluation metric, Normalized Discounted Cumulative Gain (NDCG@k), is used.  Specifically: 
\begin{itemize}
    \item For \texttt{literal} answer types, only a single predicted type is considered that can be either correct (NDCG=1) or incorrect (NDCG=0).
    \item For \texttt{resource} answer types, a ranked list of top-$k$ ontology classes is considered and evaluated in terms of lenient NDCG@k with linear decay~\citep{Balog:2012:HTT}.  The gain for a given predicted type is 0 if it is not on the same path with any of the gold types, and otherwise it is $1-d(t,t_q)/h$, where $d(t,t_q)$ is the distance between the predicted type and the closest matching gold type in the type hierarchy, with $h$ being the maximum depth of the type hierarchy.
\end{itemize}
In case of Wikidata, the type hierarchy is rather flat. Therefore, type prediction is evaluated using a binary notion of relevance, with Mean Reciprocal Rank (MRR) as the metric.

We report results on the training dataset, using 5-fold cross-validation.  For our official submissions, we also report the performance on the test set.

\subsection{Results}

\subsubsection{Category Classification}
It can be seen from the results in Table~\ref{tab:cat} that both feature-based and neural approaches perform quite well for category classification. BERT has a slight advantage over SVM. We hypothesize that due to the clear patterns which the models can learn, the high-level category classification is a fairly easy task and hence the high accuracy scores. However, most mistakes occur for the \texttt{resource} class, which is the majority class in both datasets.

\subsubsection{Type Prediction}
Since different metrics are used for DBpedia and Wikidata, we report results on the two datasets separately, in Tables~\ref{tab:dbpedia_ndcg} and~\ref{tab:wikidata_mrr}, respectively.  Recall, that (stage-two) type prediction is applied on top of (stage-one) category classification (SVM or BERT) and is only carried out when the predicted category is \texttt{resource}.  We thus prefix the method names in the result tables with SVM- or BERT- to indicate how category classification was performed.

On DBpedia (Table~\ref{tab:dbpedia_ndcg}), XBERT clearly outperforms the IR approaches. We attribute this to the fact that XBERT is tailored for XMC problem which can deal with large number of types and sparsity with tail resource types. The slight difference between SVM-XBERT and BERT-XBERT is due to the mistakes made by SVM in category classification. 
Given the large advantage of XBERT over the IR approaches, our official submissions on Wikidata (Table~\ref{tab:wikidata_mrr}) only considered the former.
It should nevertheless be noted that the IR approaches are unsupervised methods that do not need any training data. Supervised alternatives have shown to perform significantly better~\citep{Garigliotti:2017:TTI}. We leave that comparison to future work.

\begin{table*}[t!]
\caption{Category classification results, measured in terms of Accuracy. Best scores for each dataset are in boldface.}
\label{tab:cat}
\centering
\begin{tabular}{l@{~~~}l@{~~}r@{~~}r} 
\toprule
\textbf{Dataset} & \textbf{Method} & \textbf{Train} & \textbf{Test} \\
\midrule
 DBpedia & SVM &  0.958 & 0.964\\
 & BERT & \textbf{0.970} & \textbf{0.977}\\
 \midrule
 Wikidata & SVM & 0.956 & 0.960\\
 & BERT & \textbf{0.964} & \textbf{0.970}\\
\bottomrule
\end{tabular}

\end{table*}

\begin{table}[t!]
\centering
\caption{Type prediction results on DBpedia. Best scores are in boldface.}
\label{tab:dbpedia_ndcg}
\begin{tabular}{lrrrr}
\toprule
\multirow{2}{*}{\textbf{Method}}
& \multicolumn{2}{c}{\textbf{Train}}
& \multicolumn{2}{c}{\textbf{Test}} \\
   &  ~~\textbf{NDCG@5} & ~~\textbf{NDCG@10} & ~~\textbf{NDCG@5} & ~~\textbf{NDCG@10} \\
\midrule
BERT-IR/TC  & 0.492 & 0.509 & - & - \\
BERT-IR/EC  & 0.483 & 0.503 & - & - \\
SVM-XBERT  & 0.773 & 0.744 & 0.790 & 0.778\\
BERT-XBERT  & \textbf{0.776} & \textbf{0.747} & \textbf{0.804} & \textbf{0.793} \\

\bottomrule
\end{tabular}
\end{table}

\begin{table*}[t!]
\caption{Type prediction results on Wikidata (measured in terms of MRR). Best scores are in boldface.}
\label{tab:wikidata_mrr}
\centering
\begin{tabular}{lrr}
    \toprule
    \textbf{Method}  & \textbf{Train data} & \textbf{Test data} \\
    \midrule
    SVM-XBERT  & 0.66 & 0.67 \\
    BERT-XBERT  & \textbf{0.67} & \textbf{0.68} \\
    \bottomrule
\end{tabular}
\end{table*}

\section{Error Analysis}
\label{sec:analysis}

\begin{table*}[ht!]
\caption{Top-10 mistakes in type prediction by the BERT-XBERT model for DBpedia and Wikidata.}
\label{tab:top_mistakes}
\centering
\begin{tabular}{|l|r|r||p{2.5cm}|r|r|} 
\hline
\textbf{DBpedia Type} & \textbf{\#Total} & \textbf{\#Errors} & \textbf{Wikidata Type} & \textbf{\#Total} & \textbf{\#Errors} \\

\hline
\hline \texttt{dbo:Person} & 2713 & 184 & \texttt{natural person} & 1901 & 214 \\
\hline \texttt{dbo:Country} & 751 & 86 & \texttt{omnivore}  & 1901 & 214 \\
\hline \texttt{dbo:State} &  490 & 59 & \texttt{person} & 1963 & 204 \\
\hline \texttt{dbo:Organisation} & 1401 & 45 & \texttt{political territorial entity} & 570 & 120\\
\hline \texttt{dbo:Media} & 188 & 39 & \texttt{country}& 685  & 119\\
\hline \texttt{dbo:Company} & 440 & 38 & \texttt{state} &  692 & 119\\
\hline \texttt{dbo:Activity} & 204 & 38 & \texttt{organization} & 426 & 110\\
\hline \texttt{dbo:Work} & 898 & 26 & \texttt{class} & 375 & 103\\
\hline \texttt{dbo:Band}  & 98 & 25 & \texttt{community} & 326 & 86\\
\hline \texttt{dbo:Profession} & 97 & 25 & \texttt{big city} & 235 & 82\\
\hline
\end{tabular}

\end{table*}

In this section, we analyze the errors made by the our best performing approach, BERT-XBERT. 
First, we look at resource types where most errors occur.  That is, types which are present in the gold labels but are missing from the predicted labels. Table~\ref{tab:top_mistakes} shows the top-10 errors in type prediction for DBpedia and Wikidata, together with their total instance counts.  Ideally, we would expect that the number of mistakes to be directly proportional to the total frequency of the resource type.  In DBpedia, some types such as \texttt{dbo:State}, \texttt{dbo:Activity}, \texttt{dbo:Band}, and \texttt{dbo:Profession} break this pattern. Similarly in Wikidata, \texttt{natural person}, \texttt{political territorial entity}, and \texttt{big city} are some of types with which the BERT-XBERT model struggles.


\begin{table}[ht!]
    \caption{Example questions from DBpedia (top two) and Wikidata (bottom two) with respective gold and predicted type labels (by BERT-XBERT method).}
    \label{tab:mistake_examples}    \centering
    \begin{tabular}{|p{4cm}|p{3.1cm}|p{4.8cm}|}
    \hline
    \textbf{Question} & \textbf{Gold types} & \textbf{Predicted types} \\
    \hline
    \hline
     What is the title of Kakae ?
    & \texttt{["dbo:Settlement", "dbo:Island"]}
    & \texttt{["dbo:Agent", "dbo:Media", "dbo:Organisation", "dbo:PersonFunction", "dbo:Profession", "dbo:Person", "dbo:Island", "dbo:University", "dbo:Bone", "dbo:Factory"]} \\
    \hline
    \hline
    List the destination of Novair International Airways ?
    & \texttt{["dbo:Sea", "dbo:Country", "dbo:River", "dbo:Continent"]}
    & \texttt{["dbo:Country", "dbo:Location", "dbo:Place", "dbo:PopulatedPlace", "dbo:Continent", "dbo:Airport", "dbo:MeanOfTransportation", "dbo:Aircraft", "dbo:Infrastructure", "dbo:ArchitecturalStructure"]} \\
    \hline
    which cola starts with the letter p &  \texttt{["soft drink", "trademark", "carbonated beverage", "non-alcoholic beverage", "symbol", "class", "protected name"]}   &   \texttt{["non-alcoholic beverage", "carbonated beverage", "soft drink", "trademark", "food", "long gun", "goods", "dish", "cyclic process", "tea"]}\\
    \hline
   What periodical literature does Delta Air Lines use as a moutpiece?&
    \texttt{["publication", "recurring", "intellectual work", "text", "communication medium", "serial"]}&
  \texttt{["organization", "creative work", "written work", "text", "magazine", "media enterprise", "newspaper", "communication medium", "genre", "periodical"]}\\
  \hline
    \end{tabular}
\end{table}

In Table \ref{tab:mistake_examples}, we show anecdotal examples of the mistakes made by the BERT-XBERT approach. Most of these errors are due to irrelevant types returned in the result list.  In several cases, the predicted labels do contain the the gold label but place them at lower ranks, which affects the NDCG and MRR scores.  In some cases the predicted labels are appropriate, even though they do not exactly match the gold labels.  For example, for the last question in Table~\ref{tab:mistake_examples}, \texttt{publication} is one of the gold labels, which is not predicted, but \texttt{written work} and \texttt{periodical} are still relevant among the predicted labels.  We also spotted several instances with double questions such as ``What conflict occurred in Philoctetes and who was involved?'' and questions with grammatical errors and typos.


\section{Conclusions}

In this paper, we presented our solution for the SMART Task challenge of ISWC 2020, which was the best performing approach on both datasets and tasks, across all evaluation metrics.  Our findings suggest that for coarse-grained category prediction, simple feature-based approaches are quite effective with over 95\% accuracy, while sophisticated neural Transformer architectures only improve marginally. For fine-grained type prediction, on the other hand, Transformer models for extreme multilabel classification clearly outperform retrieval-based approaches.

Our future work concerns an in-depth analysis of the results on DBpedia vs. Wikidata, to understand the differences and modeling requirements for small and hierarchical (DBpedia) vs. large and shallow (Wikidata) type taxonomies.

\newpage

\providecommand{\bibfont}{\footnotesize}
\bibliographystyle{abbrvnat}
\bibliography{references}
\end{document}